\documentclass[sigconf]{acmart}

\usepackage{amsmath,multirow}
\usepackage{caption,subfigure}
\usepackage{balance}

\AtBeginDocument{%
  }

\setcopyright{acmcopyright}
\copyrightyear{2023}
\acmYear{2023}
\setcopyright{acmlicensed}\acmConference[MM '23]{Proceedings of the 31st ACM International Conference on Multimedia}{October 29-November 3, 2023}{Ottawa, ON, Canada}
\acmBooktitle{Proceedings of the 31st ACM International Conference on Multimedia (MM '23), October 29-November 3, 2023, Ottawa, ON, Canada}
\acmPrice{15.00}
\acmDOI{10.1145/3581783.3613753}
\acmISBN{979-8-4007-0108-5/23/10}

\begin{document}

\title{DAE-Talker: High Fidelity Speech-Driven Talking Face Generation with Diffusion Autoencoder}


\author{Chenpeng Du}
\authornote{Work done as an intern at Microsoft Research Asia.}
\affiliation{%
  \institution{X-LANCE Lab,\\Shanghai Jiao Tong University}
  \city{Shanghai}
  \country{China}}
\email{duchenpeng@sjtu.edu.cn}

\author{Qi Chen}
\affiliation{%
  \institution{X-LANCE Lab,\\Shanghai Jiao Tong University}
  \city{Shanghai}
  \country{China}}
\email{cq1073554383@sjtu.edu.cn}

\author{Tianyu He}
\affiliation{%
  \institution{Mircrosoft Research Asia}
  \city{Beijing}
  \country{China}}
\email{tianyuhe@microsoft.com}

\author{Xu Tan}
\authornote{Xu Tan and Kai Yu are the corresponding authors.}
\affiliation{%
  \institution{Mircrosoft Research Asia}
  \city{Beijing}
  \country{China}}
\email{xuta@microsoft.com}

\author{Xie Chen}
\affiliation{%
  \institution{X-LANCE Lab,\\Shanghai Jiao Tong University}
  \city{Shanghai}
  \country{China}}
\email{chenxie95@sjtu.edu.cn}

\author{Kai Yu}
\authornotemark[2]
\affiliation{%
  \institution{X-LANCE Lab,\\Shanghai Jiao Tong University}
  \city{Shanghai}
  \country{China}}
\email{kai.yu@sjtu.edu.cn}

\author{Sheng Zhao}
\affiliation{%
  \institution{Mircrosoft Cloud+AI}
  \city{Beijing}
  \country{China}}
\email{szhao@microsoft.com}

\author{Jiang Bian}
\affiliation{%
  \institution{Mircrosoft Research Asia}
  \city{Beijing}
  \country{China}}
\email{jiabia@microsoft.com}


\def\authors{Chenpng Du, Qi Chen, Tianyu He, Xu Tan, Xie Chen, Kai Yu, Sheng Zhao, Jiang Bian}


\renewcommand{\shortauthors}{Chenpeng Du et al.}

\begin{abstract}
While recent research has made significant progress in speech-driven talking face generation, the quality of the generated video still lags behind that of real recordings. One reason for this is the use of handcrafted intermediate representations like facial landmarks and 3DMM coefficients, which are designed based on human knowledge and are insufficient to precisely describe facial movements. Additionally, these methods require an external pretrained model for extracting these representations, whose performance sets an upper bound on talking face generation. To address these limitations, we propose a novel method called DAE-Talker that leverages data-driven latent representations obtained from a diffusion autoencoder (DAE). DAE contains an image encoder that encodes an image into a latent vector and a DDIM image decoder that reconstructs the image from it. We train our DAE on talking face video frames and then extract their latent representations as the training target for a Conformer-based speech2latent model. During inference, DAE-Talker first predicts the latents from speech and then generates the video frames with the image decoder in DAE from the predicted latents.
This allows DAE-Talker to synthesize full video frames and produce natural head movements that align with the content of speech, rather than relying on a predetermined head pose from a template video. We also introduce pose modelling in speech2latent for pose controllability. Additionally, we propose a novel method for generating continuous video frames with the DDIM image decoder trained on individual frames, eliminating the need for modelling the joint distribution of consecutive frames directly. Our experiments show that DAE-Talker outperforms existing popular methods in lip-sync, video fidelity, and pose naturalness. We also conduct ablation studies to analyze the effectiveness of the proposed techniques and demonstrate the pose controllability of DAE-Talker. 
The synthetic results are available at \href{https://daetalker.github.io}{https://daetalker.github.io}.
\end{abstract}





\maketitle

\section{Introduction}

Generating high-fidelity videos of talking faces that synchronize with input speech is a challenging and crucial task. Researchers have proposed various approaches to tackle this problem, broadly categorized as 2D-based and 3D-based methods. The former typically uses facial landmarks as an intermediate representation. The system first predicts these landmarks from the speech input and then generates corresponding video frame. For instance, \cite{synobama} predicts only mouth-related landmarks and textures, while \cite{makeittalk} predicts all facial landmarks and uses them to warp a static image. Some methods, such as \cite{wav2lip,pretrainedG}, directly optimize their models using a generative adversarial network (GAN), without any intermediate representation. In contrast, 3D-based methods use 3D morphable models (3DMM) coefficients \cite{nvp,ranyi3d,lipsync3d}, which are obtained from a pretrained 3D face reconstruction model, as their intermediate representation. Recently, some researchers have applied the neural radiance field (NeRF) to this task, as reported in \cite{adnerf,dfrf}. NeRF generates the final image by volume rendering.

Despite recent advancements in speech-driven talking face generation, the fidelity of the generated video still lags behind that of real recordings. One reason for this is the use of handcrafted intermediate representations, such as facial landmarks and 3DMM coefficients, which are designed based on human knowledge and are insufficient to precisely describe facial movement. Additionally, these methods necessitate an external pretrained model to extract these representations, which may introduce additional errors. The performance of the representation extractor sets an upper bound on talking face generation.

In this paper, we introduce a novel system named DAE-Talker for generating speech-driven talking face. Our method employs data-driven latent representations from a diffusion autoencoder (DAE) \cite{diffae}. DAE contains an image encoder that encodes an image into a latent vector and an image decoder based on denoising diffusion implicit model (DDIM) that reconstructs the image from the latent vector. We train the DAE on talking face video frames and then extract their intermediate latent representations as the training target for a Conformer-based speech2latent model. Different from using a pretrained model for extracting the traditional handcrafted representations, the image encoder in DAE is jointly optimized with the DDIM image decoder in a data-driven manner to achieve a better reconstruction performance. During inference, DAE-Talker first predicts the latents from speech and then generates the video frames with the image decoder in DAE from the predicted latents.
This allows DAE-Talker to synthesize full video frames and produce natural head movements that align with the content of speech, rather than relying on a predetermined head pose from a template video. Moreover, we model the head pose explicitly in speech2latent for the pose controllability. Hence, DAE-Talker can either generate natural head movements based on the input speech or control the head pose with a specified one.

It is worth noting that the DAE is trained on individual video frames, without taking into account the correlation between consecutive frames. In this work, we introduces a novel method that employs the DDIM image decoder to generate continuous video frames. The denoising process of DDIM behaves as an ODE solver \cite{ddim}, resulting in a continuous deterministic trajectory. Rather than sampling various Gaussian noises to produce different frames, we propose to utilize a shared Gaussian noise $x_T$ to generate all video frames. This allows the DDIM's denoising process to initiate from a fixed point. Additionally, our speech2latent model is trained on sequence-level, with a global and local context-aware architecture, leading to a continuous predicted latent sequence. This continuity enables the ODE trajectory to change smoothly along different frames. As a result, we can produce continuous video frames with the DDIM image decoder, obviating the need to directly model the joint distribution of consecutive video frames.

 Our experiments show that DAE-Talker outperforms existing popular methods in lip-sync, video fidelity, and pose naturalness. We also conduct ablation studies to analyze the effectiveness of the proposed techniques and demonstrate the pose controllability of DAE-Talker.
The main contributions of this work are as follows:
\begin{itemize}
    \item We propose a novel approach for generating talking faces, which involves utilizing data-driven latent representations of a diffusion autoencoder. Our method outperforms existing popular methods in terms of lip-sync accuracy, video fidelity, and pose naturalness.
    \item DAE-Talker is able to synthesize full video frames and produce natural head movements that align with the content of speech, rather than relying on a predetermined head pose from a template video. 
    \item To generate continuous video frames, we propose to start the denoising process of the DDIM image decoder from a shared Gaussian noise $x_T$ for all video frames and produce smoothly changed ODE trajectory by continuous latent sequence predicted by the speech2latent model. 
\end{itemize}

The remainder of this paper is structured as follows. In Section 2, we conduct a review of the related work on the diffusion model and speech-driven talking face generation. Section 3 presents the proposed DAE-Talker and the theoretical validity of our approach for generating continuous video frames. In Section 4, we demonstrate and analyze the results of our experiments. Finally, we conclude this paper in Section 5.

\begin{figure*}[h]
\centering
\subfigure[\textbf{Training Stage 1:} Diffusion autoencoder.]{
\includegraphics[page=2,width=0.27\linewidth,trim=12.1cm 4.1cm 11.8cm 3cm,clip=true]{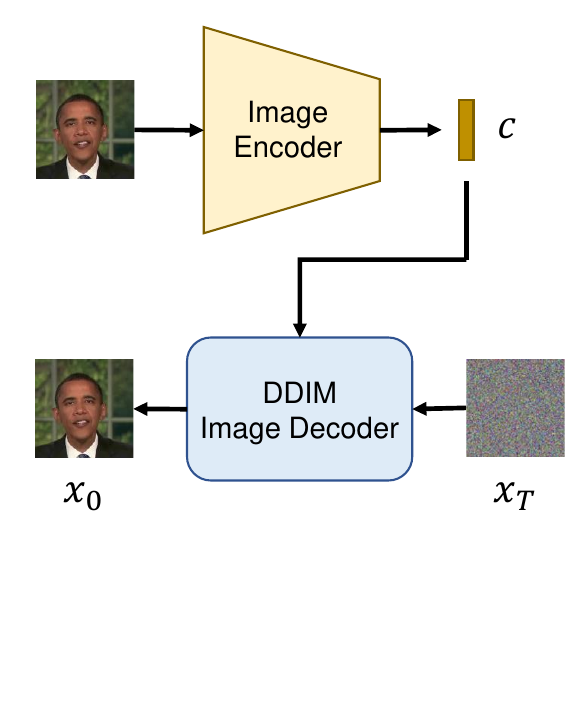}
\label{fig:diffae}
}
\subfigure[\textbf{Training Stage 2:} Speech2latent.]{
\includegraphics[page=3,width=0.68\linewidth,trim=4.1cm 2.8cm 4.9cm 2.2cm,clip=true]{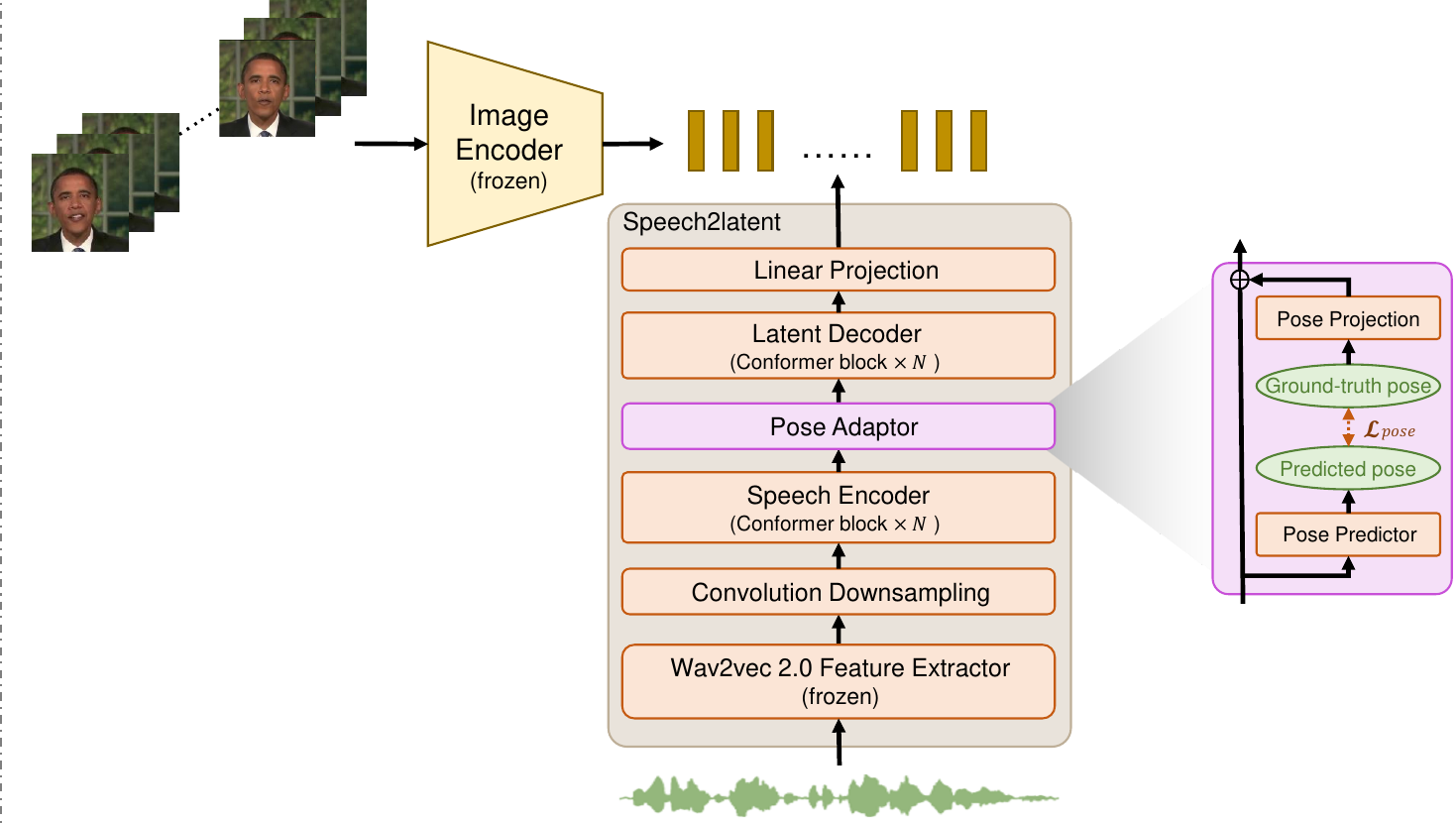}
\label{fig:s2l}
}
\caption{Two-stage training process of DAE-Talker. In stage 1, we first train a diffusion autoencoder on video frames for extracting latent representation. In stage 2, we train a speech2latent model for predicting latent from speech.}
\label{fig:training_pipeline}
\end{figure*}

\section{Related work}
\subsection{Diffusion model}

Recent advancements in image generation have been made by denoising diffusion probabilistic models (DDPM) \cite{ddpm,diffusion15} and score-based models \cite{scoremodel}, surpassing GAN-based models \cite{beatgan,beat_gan2} in both fidelity and diversity. DDPM is a parameterized Markov chain that incrementally adds Gaussian noise to images and uses a deep neural network to reconstruct the original image from the noise step by step. 
DDIM \cite{ddim} utilizes a class of non-Markovian diffusion processes, leading to accelerated sampling. An autoencoder based on DDIM was proposed by \cite{diffae} for extracting meaningful representations in face generation. Additionally, \cite{video_diffae} disentangled face identity and motion using a diffusion autoencoder.

The diffusion model has also been applied to video generation. For instance, \cite{video_diffusion} utilizes DDPM to model the joint distribution of multiple consecutive video frames. To improve the generation speed, \cite{latent_video_diffusion} proposes to model the video in a low dimensional latent space. \cite{latent_video_diffusion_arbitrarylen} further extends this approach to generate video of arbitrary length with the latent video diffusion model. Another successful method for high-resolution video generation is proposed in \cite{video_diffusion_cascade}, where the authors cascade diffusion models and super-resolution models. Despite these advancements, modeling the joint distribution of consecutive video frames often results in large model sizes and slow convergence speed. Rather than relying on the above video diffusion model, DAE-Talker generates continuous video frames using a DDIM image decoder trained on independent video frames. The correlation between consecutive frames is instead modeled by speech2latent, which is trained at sequence-level.

\subsection{Speech-driven talking face generation}

Speech-driven talking face generation has been extensively explored in recent literature. \cite{taylor2017deep} uses a sliding window predictor that learns arbitrary nonlinear mappings from phoneme label input sequences to mouth movements in a way that accurately captures natural motion and visual coarticulation effects. \cite{karras2017audio} predicts the 3D vertex coordinates of a face model from speech. 

GAN is one of the most popular tools for talking face generation. \cite{vougioukas2020realistic} introduces temporal GAN that uses 3 discriminators focused on achieving detailed frames, audio-visual synchronization, and realistic expressions. Wav2lip \cite{wav2lip} proposes to use a pretrained lip-sync expert model as the discriminator. SPACE \cite{spacex} leverages latent representation of a pretrained face-vid2vid model but still relies on facial landmark prediction. 

Neural radiance field (NeRF) is another type of talking face model. AD-NeRF \cite{adnerf} models the head and torso with two separate NeRFs, conditioning the prediction on the input speech. \cite{yao2022dfa} disentangle head poses from lip movements and sample them from a probabilistic model. DFRF \cite{dfrf} proposes dynamic facial radiance fields for few-shot talking head synthesis, generalizing to an unseen identity with few training data. However, NeRF-based methods have shown limitations in modeling the dynamic radiance field with uncertainty, sometimes resulting in blurry torso.

Our concurrent work Diffused Head \cite{diffusehead} proposes a similar idea to us, using diffusion models for speech-driven talking face generation, but they model the correlation between video frames in an auto-regressive manner, which limits the inference speed. In contrast, DAE-Talker proposes a novel approach to generate all video frames in parallel while maintaining continuity of consecutive frames.

\section{DAE-Talker}

In this paper, we propose DAE-Talker, a novel system for generating speech-driven talking faces that utilizes data-driven latent representations from a diffusion autoencoder (DAE). The training process involves two stages, as shown in Figure \ref{fig:training_pipeline}. We begin by training a DAE on video frames of talking faces, then we extract the latent representations and use them as the training target for a Conformer-based speech2latent model. During inference, we first predicts the latents from speech and then generates the video frames with the image decoder in DAE from the predicted latents. In this section, we discuss the training stages and the inference process.

\subsection{Latent extraction from diffusion autoencoder}

As illustrated in Figure \ref{fig:diffae}, DAE contains an image encoder that encodes an image into a latent vector $c$ and a DDIM image decoder that reconstructs the image from it. 
Different from DDPM, DDIM \cite{ddim} defines a group of non-Markovian random processes by
\begin{equation}
q(x_{1:T} | x_0) := q(x_T | x_0) \prod_{t=2}^{T} q(x_{t-1} | x_{t}, x_0)
\end{equation}
where
\begin{equation}
q(x_{T} | x_0) = \mathcal{N}(\sqrt{\alpha_T} x_0, (1 - \alpha_T) \mathbf{I})
\end{equation}
and
\begin{equation}
\begin{aligned}
& q(x_{t-1} | x_t, x_0) \\
= & \mathcal{N}\left(\sqrt{\alpha_{t-1}} x_{0} + \sqrt{1 - \alpha_{t-1} - \sigma^2_t} \cdot {\frac{x_{t}  - \sqrt{\alpha_{t}} x_0}{\sqrt{1 - \alpha_{t}}}}, \sigma_t^2 \mathbf{I} \right).
\end{aligned}
\end{equation}
Based on the definitions given above, we can derive that $\mathcal{N}(x_t|x_0)$ in DDIM takes on the same form as it does in DDPM. As a result, the training process for DDIM is identical to that of DDPM. The key difference lies in the decoding process where the value of $\sigma_t$ is set to 0 in DDIM, where we have
\begin{equation}
\begin{aligned}
x_{t-1} = & \sqrt{\alpha_{t-1}} x_0 + \sqrt{1 - \alpha_{t-1}} \cdot \frac{x_{t}  - \sqrt{\alpha_{t}} x_0}{\sqrt{1 - \alpha_{t}}} \\ 
= & \sqrt{\alpha_{t-1}} \left(\frac{x_t - \sqrt{1 - \alpha_t} \cdot \epsilon_\theta^{(t)}(x_t, c)}{\sqrt{\alpha_t}}\right) \\
& + \sqrt{1 - \alpha_{t-1}} \cdot \epsilon_\theta^{(t)}(x_t, c).
\end{aligned}
\label{eq:diff}
\end{equation}
Therefore, the decoding process of DDIM is a deterministic process rather than a stochastic process, which allows us to control the generated image with only the initial point $x_T$ and the latent representation $c$.

We train the DAE on talking face video frames. After that, we extract the latent representations $c$ for all video frames, which are then used as the training target for the speech2latent model.

\subsection{Speech2latent: Latent prediction from speech}

In order to generate lip-sync video from speech, we train a model named speech2latent to predict the corresponding latent representations from the speech, as shown in Figure \ref{fig:s2l}.

\subsubsection{Deep acoustic feature}
Deep acoustic features extracted by neural networks have been found to be superior to traditional acoustic features such as MFCC and mel-spectrogram in both speech recognition \cite{vqw2v,w2v2} and speech synthesis \cite{vqtts}. These types of neural networks are typically trained with only speech data in a self-supervised manner. Recently, they are also leveraged in talking face generation models \cite{faceformer}. Therefore, we leverage the deep acoustic feature extracted by pretrained wav2vec 2.0 \cite{w2v2} in this work.

\subsubsection{Local and global context-aware architecture}
\label{sec:speech2latent}

Latent prediction from speech is a sequence-to-sequence task, where the two sequences are monotonically aligned. Accordingly, we build our speech2latent model based on the Conformer architecture, which utilizes a convolution layer, a speech encoder, a pose adaptor, a latent decoder and a linear projection layer. Because speech has a higher frame rate than video, we set the stride size of the first convolution layer to the ratio of the two frame rates to ensure appropriate downsampling. Both the speech encoder and latent decoder consist of $N$ Conformer blocks, which is a variant of Transformer block that incorporates a convolution layer after each self-attention layer. This combination of the two layers allows for capturing both local and global context information. During training, we calculate the mean square loss $\mathcal{L}_{latent}$ between the predicted latent and the target one extracted from the ground-truth video frames.


\subsubsection{Pose modelling}
\label{sec:pose}

Note that latent prediction from speech is a one-to-many mapping. When speaking a sentence, there are multiple possible poses that can be considered correct. To address this, we propose to condition the latent prediction on the extracted head pose during training. This approach enhances the latent prediction by incorporating additional condition information and, as a result, alleviate the problem. Furthermore, it provides us with the ability to control the head pose in talking-face generation.

During training, we utilize a pose predictor to estimate the head poses based on the speech input. Also, the ground-truth head poses are projected and added to the speech encoder output to predict the latent sequence.
During inference, we can use either the predicted head poses or user-specified ones to predict the latent sequence.
The overall training criterion of speech2latent can be written as
\begin{equation}
\mathcal{L}_{speech2latent} = \mathcal{L}_{latent} + \alpha \mathcal{L}_{pose}
\end{equation}
where $\mathcal{L}_{pose}$ is the mean square loss of pose prediction and $\alpha$ is the relative weight between the two terms.

\subsubsection{Pseudo-sentence sampling for data augmentation}

There are two main methods for processing data in talking face systems. The first method predicts the target solely based on the corresponding speech frame and its consecutive frames, such as in AD-NERF, without considering long-term context. The second method divides the speech into complete sentences, such as in FaceFormer \cite{faceformer}, and trains the model on the sequence level. However, when the available data is limited to just a few minutes, the number of sentences available for training is quite small. This makes the second method more susceptible to overfitting compared to the first method that has more training samples available.

In this work, we introduces a novel data augmentation technique called pseudo-sentence sampling, which offers the benefits of both worlds. Instead of dividing the data into fixed real sentences, we randomly select a starting point and generate a pseudo-sentence by cutting it at a random length between 5 to 20 seconds for each training sample. This approach enables us to obtain sufficient training samples while maintaining sequence-level training.

\begin{figure}[t]
\centering
\includegraphics[page=1,width=0.95\linewidth,trim=11cm 5.5cm 11.5cm 3.3cm,clip=true]{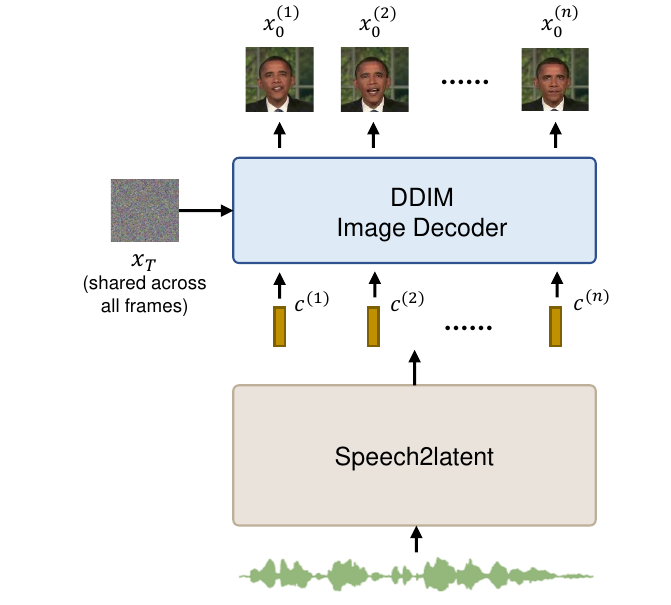}
\caption{The inference process of DAE-Talker. The Gaussian noise $x_T$ is shared across all frames for the continuity of video frames.}
\label{fig:inference}
\end{figure}

\begin{figure*}[h]
\centering
\includegraphics[page=4,width=0.9\linewidth,trim=5.9cm 2cm 2.2cm 1cm,clip=true]{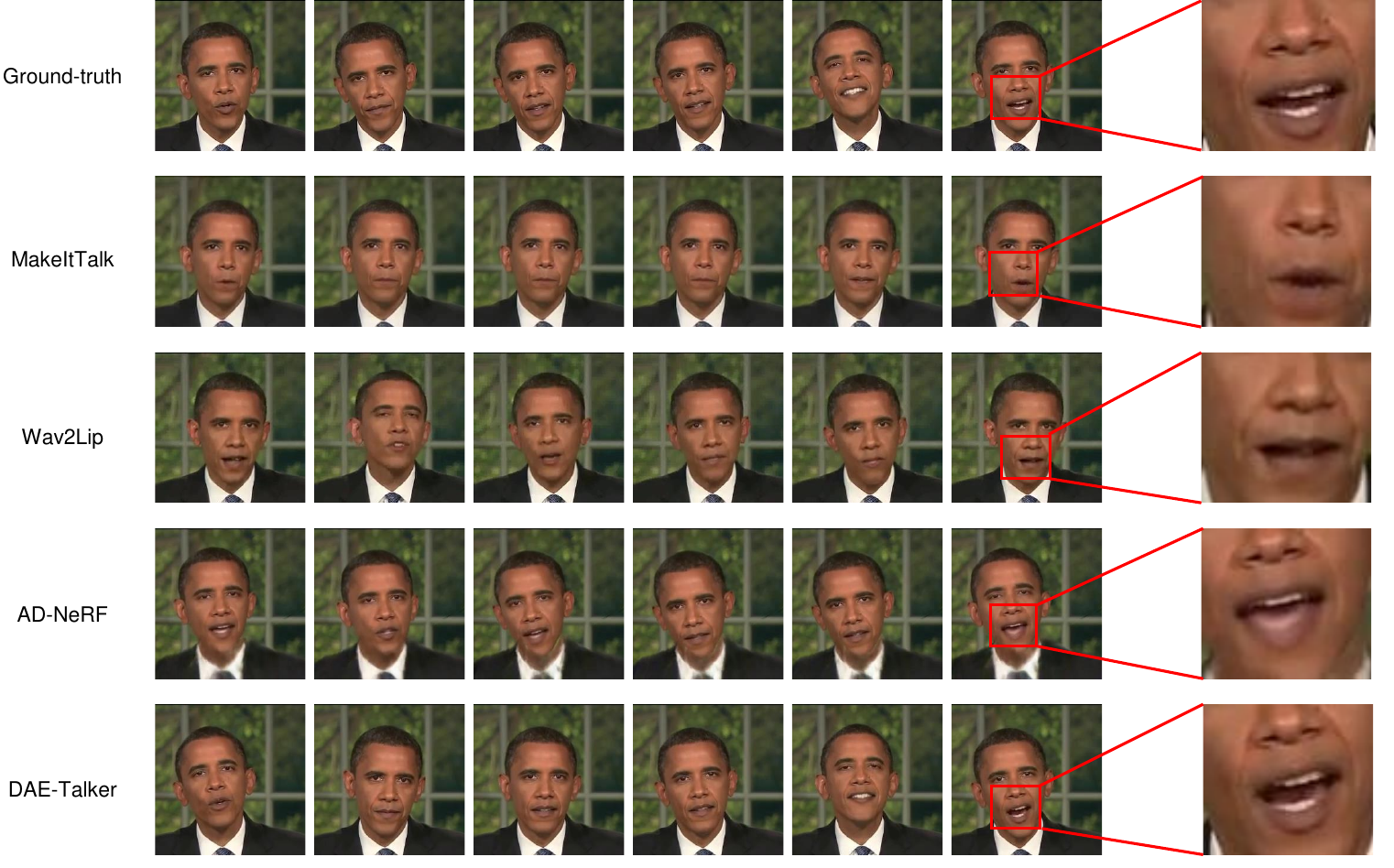}
\caption{Samples of synthetic results from DAE-Talker and the baseline methods.}
\label{fig:demo}
\end{figure*}

\subsection{Frame-wise conditioned video generation with DDIM image decoder}

Different from generating a large number of images independently, video generation requires considering the correlation between consecutive frames to ensure that the changes among frames are smooth. In this section, we discuss how to generate continuous video frames by using the DDIM image decoder.

 According to Equation \ref{eq:diff}, the denoising process of DDIM is a deterministic process, which is equivalent to an Euler solver to an ordinary differential equation (ODE). Euler method is a numerical method for solving an ODE given an initial point by taking a large amount of discrete infinitesimal steps along the direction of the gradient. Note that the direction field of the ODE is continuous, the inference process simulates the trajectory for solving the ODE which is a continuous curve. Therefore, we can approximate $x_0$, namely the ending point of the curve, as a continuous function of $x_T$ and $c$, that is
\begin{equation}
x_0 = f(x_T, c).
\end{equation}
Rather than sampling various Gaussian noises to produce different frames, we propose to utilize a shared Gaussian noise $x_T$ to generate all video frames. This provides a consistent starting point for the denoising process. Additionally, by the definition of continuous function, continuous variations in the input result in continuous changes in its output. As a result, as the latent representation $c$ changes continuously, the generated data $x_0$ also changes continuously. This statement can be formalized as
\begin{equation}
\lim_{\Delta c\rightarrow 0} f(x_T, c+\Delta c) = x_0.
\end{equation} 

Therefore, we can use a shared Gaussian noise $x_T$ and a continuous latent sequence to generate continuous video frames with the DDIM decoder, as illustrated in Figure \ref{fig:inference}. The global and local context-aware architecture of speech2latent, as explained in Section \ref{sec:speech2latent}, ensures the continuity of the latent sequence. This approach can also be applied to any frame-wise conditioned video generation task. Instead of relying on a large video diffusion model that explicitly models the correlation between consecutive frames, it suffices to use a DDIM image decoder to generate such videos.

\section{Experiments}

\subsection{Setup}

\subsubsection{Dataset}
Our dataset consists of a 15-minute video of Obama's address, which is the same one used in AD-NeRF \cite{adnerf}. The video has a resolution of $256 \times 256$ and is recorded at a frame rate of 25 fps. The audio is sampled at a rate of 16000Hz. We divide the video into two parts, using the first 12 minutes for training and the remaining 3 minutes for testing (referred to as test set A). In addition, we use another Obama's address from the Synthesizing Obama \cite{synobama} dataset as test set B. To assess our model's performance on external speakers, we also introduce LibriTTS \cite{libritts} dataset that comprises audiobook readings as the test set C.

 \begin{table*}[t]
\caption{Evaluations on lip-sync with different talking face models.}
\label{tab:talking_face_eval}
\centering
\begin{tabular}{c|ccc|ccc|cc}
\hline\hline
\multirow{2}{*}{\textbf{Model}}   &   \multicolumn{3}{c|}{\textbf{LSE-D} $\downarrow$}   &   \multicolumn{3}{c|}{\textbf{LSE-C} $\uparrow$}  &   \multicolumn{2}{c}{\textbf{LMD} $\downarrow$}  \\  \cline{2-9}
& Test set A & Test set B & Test set C & Test set A & Test set B & Test set C & Test set A & Test set B \\  \hline \hline
GT & 6.59 &  7.12  &  -  & 8.43  &  8.27 & - &  0  &  0  \\ \hline
MakeItTalk & 8.51 &  9.08  &  9.19 & 5.37 & 5.88 & 5.34 &  3.48 & 3.40 \\ 
Wav2Lip & 8.89 &  7.83  &  \textbf{7.58}  & 6.49 & 7.81 & 7.58 & 2.34 &  2.60 \\ 
AD-NeRF & 7.80 &  8.06  & 8.13  & 6.90 &  7.51  & 7.03 &  \textbf{1.06} & 2.44 \\   \hline
DAE-Talker & \textbf{7.25} &  \textbf{6.98}  & 7.63  &  \textbf{8.05} & \textbf{9.25} & \textbf{8.12} &  1.55 &  \textbf{2.25} \\    \hline \hline
\end{tabular}
\end{table*}

\begin{figure*}[h]
\centering
\subfigure[Lip-sync.]{
\includegraphics[width=0.3\linewidth]{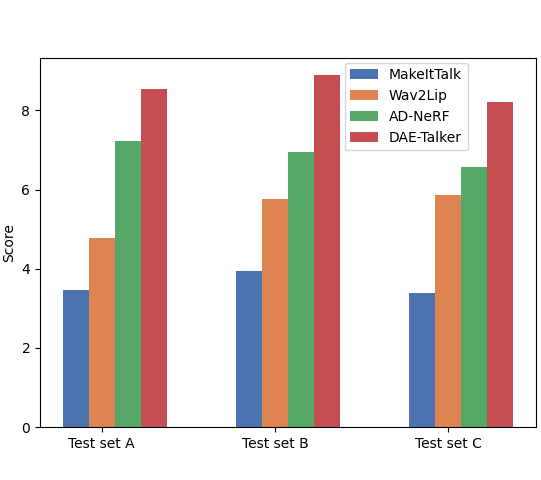}
}
\subfigure[Pose naturalness.]{
\includegraphics[width=0.3\linewidth]{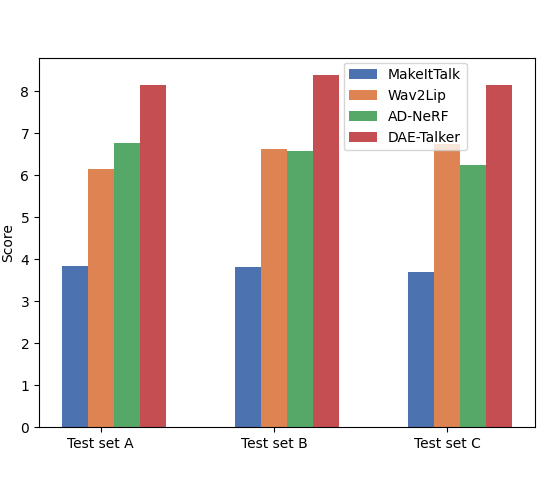}
}
\subfigure[Video fidelity.]{
\includegraphics[width=0.3\linewidth]{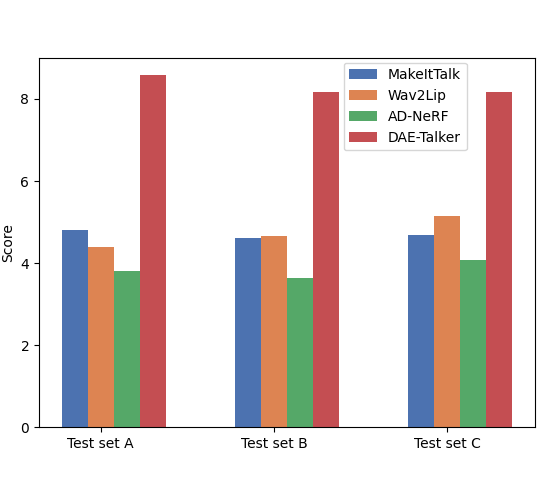}
}
\caption{Subjective evaluations on talking face models with user studies.}
\label{fig:subjective_eval}
\end{figure*}

\subsubsection{Training details}

We use Unet-based architecture for our diffusion autoencoder following \cite{diffae}. 
The dimension of the latent code is 512. The number of training diffusion steps is 1000 and we use linear scheduler for adding noise. The model is trained on L1 loss with noise prediction. The optimizer is Adam with a learning rate of 1e-4 and the batch size is set to 32. We take about 3 days to train the autoencoder for 80k steps on 8 Nvidia V100 GPUs. During inference, we use $T=100$ steps for generation.

The input of speech2latent module is the acoustic feature extracted by a pretrained wav2vec 2.0 model XLSR\footnote{https://github.com/facebookresearch/fairseq/tree/main/examples/wav2vec}. The ground-truth head pose is extracted by a python package\footnote{https://pypi.org/project/headpose} and is represented with 3 dimensional vectors including roll, pitch and yaw. Both the speech encoder and the latent decoder consists of 4 Conformer blocks with 2 attention heads. The dimension of the self attention is 256 and the kernel size of each CNN layer is 13. The pose predictor is composed of two 1D convolutional layers with kernel size of 3 and channel size of 384, each of which followed by ReLU activation, layer normalization and dropout. The optimizer of speech2latent is also Adam with a learning rate of 1e-4 and the batch size is set to 16.

\subsection{Experimental results}

In this section, we evaluate DAE-Talker's performance and compare it with several baselines using both objective and subjective measures. We introduce three baselines used in our experiments. The first one is MakeItTalk \cite{makeittalk}, which utilizes facial landmarks as an intermediate representation. The system predicts the positions of the landmarks from speech and generates an animated image by translating it using an image-to-image translation network. Although it controls the head pose implicitly through facial landmarks, it does not explicitly model it. Wav2Lip \cite{wav2lip}, on the other hand, uses a GAN-based model that takes a sequence of upper-half face images and speech segments as input and generates lower-half face images. It does not use an intermediate representation but relies on an external lip-sync expert. However, the head pose in Wav2Lip may be inconsistent with the speech content as it is controlled by a template video that provides upper-half face motion. Both of the two methods require extensive training data, and we use their released checkpoints in our evaluations.
In contrast, AD-NeRF \cite{adnerf} requires only a short video for training but can still generate natural results. It feeds speech into a conditional function and generates a dynamic neural radiance field, which is then rendered into video frames using volume rendering algorithm. In our experiments, we train the AD-NeRF model on our 12-minute training set using the official implementation\footnote{https://github.com/YudongGuo/AD-NeRF}.

 \begin{table*}[t]
\caption{Evaluations on lip-sync for ablation study.}
\label{tab:ablation}
\centering
\begin{tabular}{l|ccc|ccc|cc}
\hline\hline
\multirow{2}{*}{\textbf{Model}}   &   \multicolumn{3}{c|}{\textbf{LSE-D} $\downarrow$}   &   \multicolumn{3}{c|}{\textbf{LSE-C} $\uparrow$}  &   \multicolumn{2}{c}{\textbf{LMD} $\downarrow$}  \\  \cline{2-9}
& Test set A & Test set B & Test set C & Test set A & Test set B & Test set C & Test set A & Test set B \\  \hline \hline
DAE-Talker  & \textbf{7.25} &  \textbf{6.98}  & \textbf{7.63}  &  \textbf{8.05} & \textbf{9.25} & 8.12 &  \textbf{1.55} &  \textbf{2.25}  \\ \hline
w/o shared $x_T$ & 7.40 &  6.98  &  7.63 & 7.99 & 9.25 & \textbf{8.14} &  1.55 & 2.25 \\ 
w/o data aug & 7.82 &  7.55  &  8.17  & 7.42 & 8.61 & 7.47 & 1.88 &  2.33 \\ 
\hline \hline
\end{tabular}
\end{table*}

We evaluate lip-sync performance use three metrics: LSE-D, LSE-C \cite{wav2lip}, and LMD \cite{lmd}. LSE-D and LSE-C utilize a publicly available pre-trained SyncNet\footnote{https://github.com/joonson/syncnet} to measure the lip-sync error between the generated frames and the corresponding speech segment. On the other hand, LMD calculates the L2 distance between the landmarks of the generated video and the ground-truth one after normalization. It is worth noting that the audiobook dataset, test set C, does not have corresponding ground-truth videos, so LMD is only calculated on test sets A and B. The results are presented in Table \ref{tab:talking_face_eval}, where it shows that DAE-Talker outperforms all other methods with the lowest LSE-D and LMD and the highest LSE-C scores in most cases. Therefore, DAE-Talker has the best lip-sync performance among all the methods.

As we mentioned earlier, talking face generation is a one-to-many problem. For example, when speaking a sentence, there are multiple possible poses that can be considered correct. It is not reasonable to evaluate the pose naturalness and video fidelity by calculating the discrepancy between the generated and the ground-truth video. Therefore, we conduct user studies to evaluate lip-sync, video fidelity, and head movement naturalness. 15 participants rate each video on the above three aspects using a scale of 1 to 10, where a higher score indicates better results. The average scores are presented in Figure \ref{fig:subjective_eval}.
We found that the poses in Wav2Lip and AD-NeRF, which are cloned from a video template, have better naturalness than MakeItTalk, which only warps a static image. However, these poses could still be inconsistent with the speech content and are inferior to those of DAE-Talker. Video snapshots of DAE-Talker and the baseline methods are presented in Figure \ref{fig:demo}. Moreover, AD-NeRF models the head and torso separately and has better video fidelity than MakeItTalk and Wav2Lip in the head part. However, the torso part of AD-NeRF sometimes appears blurry, which is also mentioned in its paper. In contrast, DAE-Talker generates the full video frames simultaneously with the DDIM image decoder and achieves the highest scores in video fidelity. It also outperforms other methods in terms of lip-sync, which is consistent with the results of objective evaluations.

\begin{figure}[h]
\centering
\includegraphics[page=5,width=0.7\linewidth,trim=11cm 9.5cm 13cm 5.3cm,clip=true]{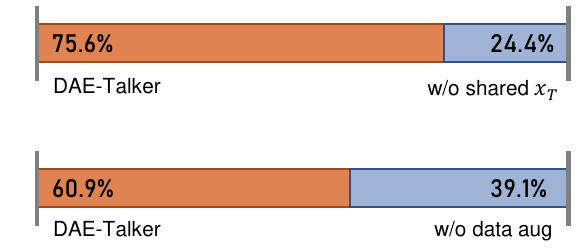}
\caption{AB preference test for ablation study.}
\label{fig:ab}
\end{figure}

\subsection{Ablation study}

We propose to generate continuous video frames with the DDIM image decoder from a fixed noise across different frames. Also, in order to model the long-term context with limited training data, we introduce pseudo-sentence sampling for data augmentation. In this section, we perform ablation studies on the two techniques. 
 
 \textbf{Ablation experiment 1: w/o shared $x_T$.} We use the same model as DAE-Talker but sample different noise $x_T$ from $\mathcal{N}(0, \mathbf{I})$ for generating different frames with the DDIM. 

 \textbf{Ablation experiment 2: w/o data aug.} We train a new speech2latent model without data augmentation with pseudo-sentence sampling, where we only cut the speech into 20 seconds segments at fixed timestamps. 

It can be observed that discontinuous frames and jiggling talking face often occur in the ablation experiment 1, while the lip-sync performance is similar to DAE-Talker.
In the ablation experiment 2, the generated video sometimes has unnatural head movement. The lip-sync performance also degrades compared with DAE-Talker.

Here, we present the results of our evaluation of two ablation experiments in Table \ref{tab:ablation} using the same objective metrics as in the previous section. The results indicate that experiment 1 has lip-sync performance similar to that of DAE-Talker, while experiment 2 is slightly worse. In addition, we conducted an AB preference test, where 15 raters are asked to choose between the videos generated by the ablation experiments and that of DAE-Talker based on video fidelity for experiment 1 and pose naturalness for experiment 2. The results are shown in Figure \ref{fig:ab}, indicating that both ablation approaches lead to worse performance compared to DAE-Talker. These results are in line with our preliminary observations.

\subsection{Analysis and discussion}

\subsubsection{Reconstruction quality from intermediate representation}

DAE-Talker leverages the diffusion autoencoder to provide latent representations of video frames. In this section, we analyze the performance of reconstructing from the latent representation. Our DAE is trained on the 12 minutes video of Obama's address and is evaluated on the test set A. The noise $x_T$ is randomly sampled from $\mathcal{N}(0, \mathbf{I})$ and the number of steps for denoising is set to $T=100$. The metrics for the evaluation are the peak signal-to-noise ratio (PSNR), the structural similarity (SSIM) \cite{ssim} and the learned perceptual image patch similarity (LPIPS) \cite{lpips}. 
The baseline in this experiment is the image generator in MakeItTalk that reconstructs the image from the facial landmarks. We use the pretrained model released by the authors in testing.
 
 \begin{table}[htbp]
\caption{Reconstruction quality from intermediate representation. ``Inter. Repr'' is the abbreviation of intermediate representation.}
\label{tab:reconstruct}
\centering
\begin{tabular}{c|c|ccc}
\hline
\textbf{Model} & \textbf{Inter. Repr.} & \textbf{PSNR}  & \textbf{SSIM}  & \textbf{LPIPS} \\ \hline
MakeItTalk   & Landmark & 20.070 & 0.692 & 0.171 \\ 
DAE-Talker   & DAE latent   & \textbf{33.702} & \textbf{0.965} & \textbf{0.012}  \\ \hline
\end{tabular}
\end{table}

Table \ref{tab:reconstruct} presents the results of our study, which demonstrate that the image reconstruction from diffusion latent outperforms the landmark-based method. 

\subsubsection{Alleviating one-to-many mapping with pose modelling}

\begin{figure}[h]
\centering
\includegraphics[width=0.9\linewidth]{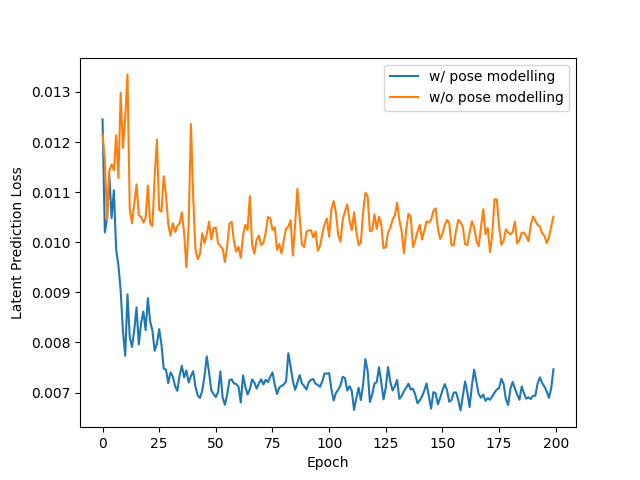}
\caption{The curves of latent prediction loss during training.}
\label{fig:latent_curve}
\end{figure}

In Section \ref{sec:pose}, we discuss the challenge of the one-to-many mapping from speech to latent, as multiple videos can correspond to the same speech input. To address this issue, we incorporate head pose modeling into our speech2latent model, conditioning the latent prediction on the head pose. In this section, we investigate the necessity of pose modeling by comparing the performance of our speech2latent model with and without the pose adaptor. Specifically, we train the models and plot their mean square error curves for latent prediction during training on the test set A. Figure \ref{fig:latent_curve} displays the results.

We can find that the mean square error of latent prediction decreases much faster and achieves a lower value when using the pose adaptor, which shows the necessity of pose modelling for alleviating the one-to-many mapping issue.

\subsubsection{Pose controllability}

DAE-Talker leverages head poses to condition the latent prediction, allowing us to utilize either the natural head movements predicted from the input speech or manually specified head poses. In this section, we assess the controllability of DAE-Talker in terms of head poses. Specifically, we synthesize lip-sync videos based on the speech data in three test sets. In the first experiment, we generate talking faces with natural head poses predicted from the input speech. In the second experiment, we manually specify three different head poses, namely towards left, right, and straight ahead. We then measure the mean square distance between the desired specified head poses and the poses extracted from the generated videos. The results are presented in Table \ref{tab:pose}.

\begin{figure}[h]
\centering
\subfigure[Natural head movement predicted from speech.]{
\includegraphics[height=1.5cm]{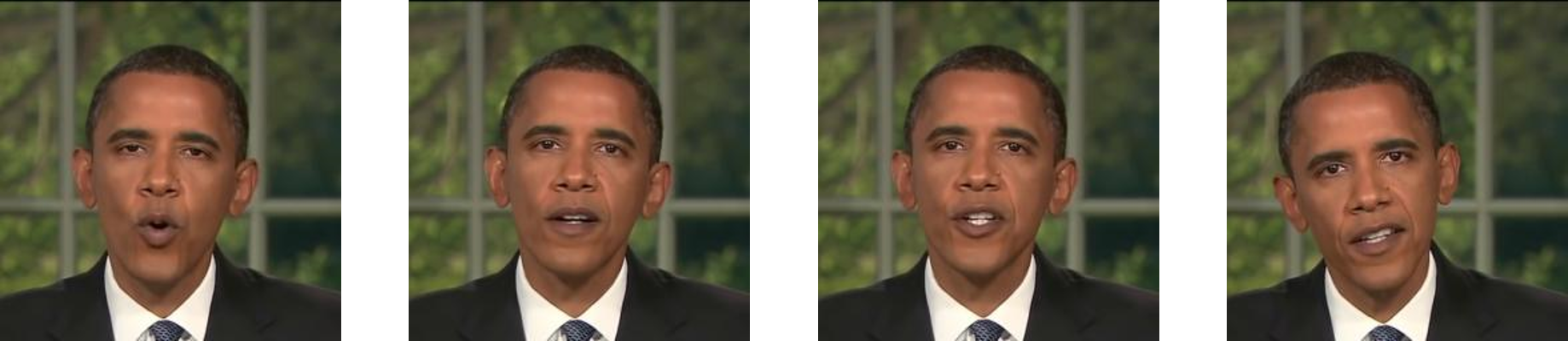}
}
\subfigure[Heading straight forward.]{
\includegraphics[height=1.5cm]{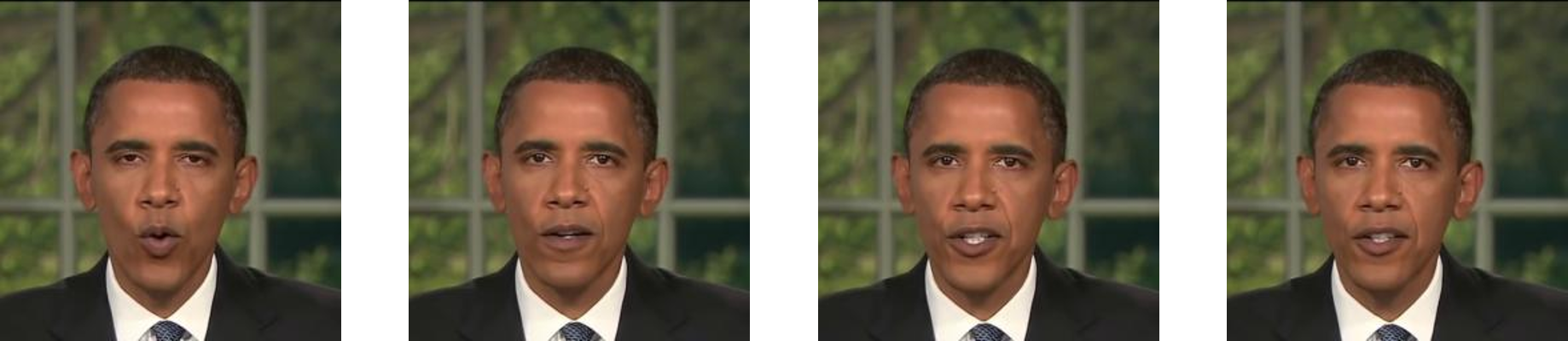}
}
\subfigure[Heading towards the left.]{
\includegraphics[height=1.5cm]{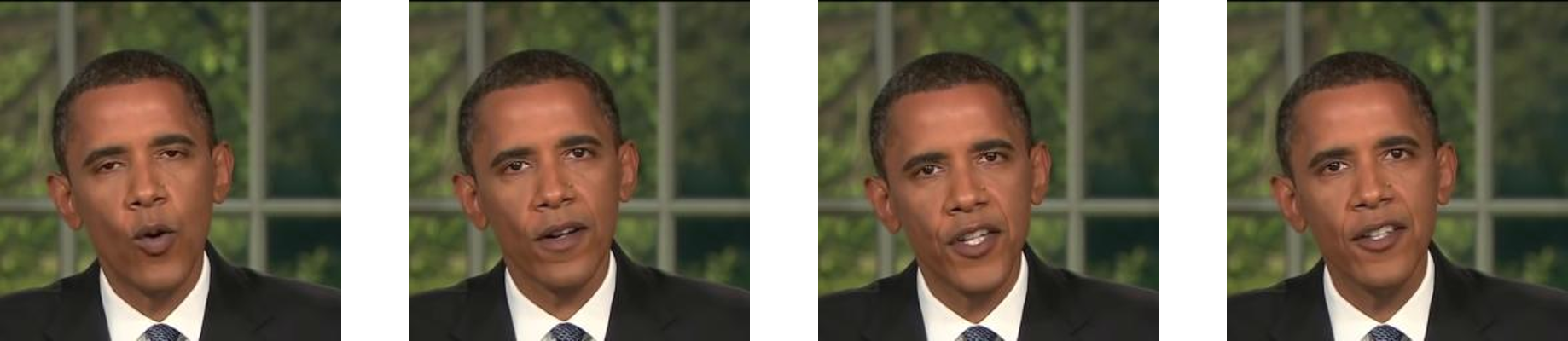}
}
\subfigure[Heading towards the right.]{
\includegraphics[height=1.5cm]{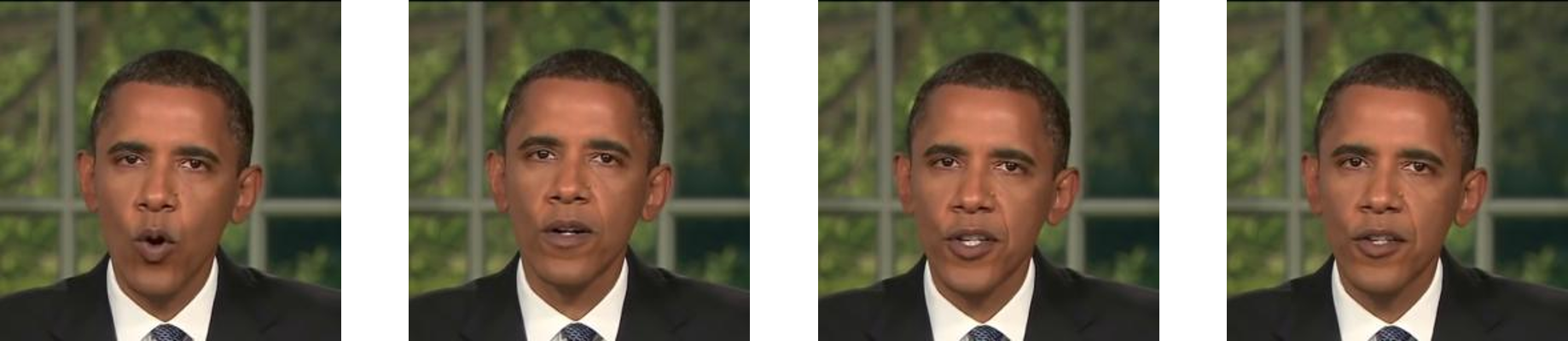}
}
\caption{Samples of pose controlled synthesis.}
\label{fig:pose_control}
\end{figure}

 \begin{table}[htbp]
\caption{The mean square distance between the specified head poses and the poses extracted from the generated video.}
\label{tab:pose}
\centering
\begin{tabular}{c|ccc}
\hline
 \textbf{Pose  Condition} & \textbf{Test set A}  & \textbf{Test set B}  & \textbf{Test set C} \\ \hline
Natural   &   113.93    &   90.84           &  76.94 \\ 
Specified   &     \textbf{7.19}     &    \textbf{7.11}     &   \textbf{6.00}  \\ \hline
\end{tabular}
\end{table}

The prediction of head poses from speech is a one-to-many mapping, which can result in a large distance between the predicted natural head poses and the specified head poses. However, the use of specified head poses can significantly reduces this distance, demonstrating the effectiveness of pose controllability.
Several snapshots of the generated videos are shown in Figure \ref{fig:pose_control}, showing the generated results with controlled head poses.

\section{Conclusion}

In this paper, we introduce a novel system named DAE-Talker for generating speech-driven talking face. Our method employs data-driven latent representations from a diffusion autoencoder that contains an image encoder that encodes an image into a latent vector and an image decoder based on DDIM that reconstructs the image from the latent vector. We train the DAE on talking face video frames and then extract their intermediate latent representations as the training target for a Conformer-based speech2latent model. 
During inference, DAE-Talker first predicts the latents from speech and then generates the video frames with the image decoder in DAE from the predicted latents. This allows DAE-Talker to synthesize full video frames and produce natural head movements that align with the content of speech, rather than relying on a predetermined head pose from a template video. Moreover, we model the head pose explicitly in speech2latent for the pose controllability. 
In our experiments, we find that our system outperforms current popular methods in lip-sync, video fidelity, and pose naturalness. Our ablative studies demonstrate the effectiveness of shared noise in video generation and pseudo-sentence sampling for data augmentation. Additionally, our results show that the data-driven latent representation from diffusion autoencoder performs better in reconstruction than the hand-crafted landmark representation. Furthermore, we discover that pose modeling can alleviate the one-to-many mapping problem in latent prediction and provide pose controllability.


\bibliographystyle{ACM-Reference-Format}
\balance
\bibliography{sample-base}










\end{document}